%% file: tree.tex
\documentclass{article}

\usepackage{times}
\usepackage{graphicx} % more modern
\usepackage{subcaption}
\usepackage{balance}

\usepackage{natbib}

\usepackage{algorithm}
\usepackage{algorithmic}

\usepackage{hyperref}
\usepackage{xcolor,colortbl}
\usepackage{amssymb}
\usepackage{amsmath}
\usepackage{tikz}
\usetikzlibrary{positioning}
\usepackage{multirow}

\usepackage [autostyle, english = american]{csquotes}

\usepackage[accepted]{icml2016-modified}

\icmltitlerunning{Learning Efficient Algorithms with Hierarchical Attentive Memory}

\newcommand{\RR}{\mathbb{R}}
\newcommand{\NN}{\mathbb{N}}

\newcommand{\emb}{\text{EMBED}}
\newcommand{\join}{\text{JOIN}}
\newcommand{\search}{\text{SEARCH}}
\newcommand{\modify}{\text{WRITE}}
\begin{document} 

\twocolumn[
\icmltitle{Learning Efficient Algorithms with Hierarchical Attentive Memory}

\icmlauthor{Marcin Andrychowicz$^\ast$}{marcina@google.com}{Google DeepMind \:\:\:\:\:\:\:\:\:\:\:\:\:\:\:\:\:\:\:\:\:\:\:}
\icmlauthor{Karol Kurach$^\ast$}{kkurach@google.com}{Google / University of Warsaw\footnote{}}
\icmlauthor{\textnormal{$^\ast$ equal contribution}}{}
\icmlkeywords{attention, machine learning, ICML}

\vskip 0.3in
]
{
  \renewcommand{\thefootnote}%
      {1}
    \footnotetext{Work done while at Google.}
}
\begin{abstract}

In this paper, we propose and investigate a novel memory architecture for neural networks called
Hierarchical Attentive Memory (HAM).
It is based on a binary tree with leaves corresponding to memory cells. This allows
HAM to perform memory access in $\Theta(\log n)$ complexity, which
is a significant improvement over the standard attention mechanism that requires
$\Theta(n)$ operations, where $n$ is the size of the memory.

We show that an LSTM network augmented with HAM can learn algorithms for
problems like merging, sorting or binary searching from pure input-output examples.
In particular, it learns
to sort $n$ numbers in time $\Theta(n \log n)$ and generalizes well to input
sequences much longer than the ones seen during the training. 
We also show that HAM %is a generic structure, which
can be trained to act
like classic data structures: a stack, a FIFO queue and a priority queue.
\end{abstract} 

\input{intro}

\input{related_work}

\input{model}

\input{experiments}

\input{comparison}
\input{conclusions}
\appendix

\input{soft}
\balance
\bibliography{tree}
\bibliographystyle{icml2016}

\end{document}

%% file: intro.tex
\section{Intro}

Deep Recurrent Neural Networks (RNNs) have recently proven to be very successful in real-word
tasks, e.g. machine translation \cite{sutskever2014sequence} and computer
vision \cite{vinyals2014show}. However, the success has been achieved
only on tasks which do not require a large memory to solve the problem, e.g.
we can translate sentences using RNNs, but we can not produce reasonable
translations of really long pieces of text, like books.

A high-capacity memory is a crucial component necessary to deal with
large-scale problems that contain plenty of long-range dependencies. Currently used
RNNs do not scale well to larger memories, e.g. the number of parameters in an
LSTM \cite{lstm} grows quadratically with the size of the network's memory. In
practice, this limits the number of used memory cells to few thousands.

It would be desirable for the size of the memory to be independent of
the number of model parameters. The first versatile and highly successful
architecture with this property was Neural Turing Machine (NTM) \cite{ntm}. The
main idea behind the NTM is to split the network into a trainable
``controller'' and an ``external'' variable-size memory.
It caused an outbreak of other neural network architectures with
external memories (see Sec.~\ref{sec:related}).

However, one aspect which has been usually neglected so far is the efficiency
of the memory access. Most of the proposed memory architectures have the
$\Theta(n)$ access complexity, where $n$ is the size of the memory. It means
that, for instance, copying a sequence of length $n$ 
requires performing $\Theta(n^2)$ operations, which is clearly unsatisfactory.

\subsection{Our contribution}

In this paper we propose a novel memory module for neural networks, called
Hierarchical Attentive Memory (HAM). The HAM module is generic and can be used
as a building block of larger neural architectures.  Its crucial property is
that it scales well with the memory size --- the memory access
requires only $\Theta(\log n)$ operations, where $n$ is the size of the memory. This complexity is
achieved by using a new attention mechanism based on a binary tree with leaves
corresponding to memory cells.  The novel attention mechanism is not only
faster than the standard one used in Deep Learning \cite{bahdanau2014}, but it
also facilities learning algorithms due to a built-in bias towards operating on
intervals.

We show that an LSTM augmented with HAM is able to learn algorithms
for tasks like merging, sorting or binary searching.  In particular, it is the first neural network, which we are aware
of, that is able to learn to sort from pure input-output examples and generalizes
well to input sequences much longer than the ones seen during the training.
Moreover, the learned sorting algorithm runs in time $\Theta(n \log n)$.
We also show that the HAM memory itself is capable of simulating different classic
memory structures: a stack, a FIFO queue and a priority queue.

%% file: related_work.tex
\section{Related work}\label{sec:related}

In this section we mention a number of recently
proposed neural architectures with an external memory, which size is independent of the number
of the model parameters.

\paragraph{Memory architectures based on attention}

Attention is a recent but already extremely successful technique in Deep Learning.
This mechanism allows networks to \emph{attend} to parts of the
(potentially preprocessed) input sequence \cite{bahdanau2014} while generating the
output sequence. It is implemented by giving the network as an auxiliary input
a linear combination of input symbols, where the weights of this linear
combination can be controlled by the network. 

Attention mechanism was used to access the memory in Neural Turing Machines (NTMs)
\citep{ntm}.  It was the first paper, that explicitly attempted to train
a computationally universal neural network and achieved encouraging results.

The Memory Network \citep{memnet1} is an early model that attempted  
to explicitly separate the memory from computation in a neural network model.
The followup work of \cite{memnet2} combined the memory network with the soft
attention mechanism, which allowed it to be trained with less supervision.
In contrast to NTMs, the memory in these models is non-writeable.

Another model without writeable memory is
the Pointer Network \citep{vinyals2015pointer}, which
is very similar to the attention model of \citet{bahdanau2014}.
Despite not having a memory, this model was able to solve
a number of difficult algorithmic problems that include the Convex Hull
and the approximate 2D Travelling Salesman Problem.

All of the architectures mentioned so far use
standard attention mechanisms to access the memory and therefore memory access
complexity scales linearly with the memory size.

\paragraph{Memory architectures based on data structures}

 %\item based on \textbf{stacks, queues}:
 Stack-Augmented Recurrent Neural Network \cite{stack-rnn} is a neural architecture combining an RNN and a differentiable stack.
 %that an LSTM augmented with a differentiable stack can learn to infer a number of algorithmic patterns.
 In another paper \cite{grefenstette2015learning}  authors consider
 extending an LSTM with a stack, a FIFO queue or a double-ended queue and show some promising results.
 The advantage of the latter model is that the presented data structures have a constant access time.
 %\item based on \textbf{pointer}:
 \paragraph{Memory architectures based on pointers}
 In two recent papers \cite{rl-ntm,zaremba2015learning} authors consider
 extending neural networks with nondifferentiable memories based on pointers
 and trained using Reinforcement Learning.
 The big advantage of these models is that they allow a constant time memory access.
 They were however only successful on relatively simple tasks.
 
 Another model, which can use a pointer-based memory is the Neural Programmer-Interpreter \cite{npi}.
 It is very interesting, because it managed to learn sub-procedures.
 Unfortunately, it requires strong supervision in the form of execution traces.% along with information, which sub-procedure should be called at every timestep.
 
 Another type of pointer-based memory was presented in Neural Random-Access Machine \cite{nram}, which
 is a neural architecture mimicking classic computers.
 
 \paragraph{Parallel memory architectures}
 %\item based on \textbf{parallel computation}:
 There are two recent memory architectures, which are especially suited for parallel computation.
 Grid-LSTM \cite{grid} is an extension of LSTM to multiple dimensions.
 Another recent model of this type is Neural GPU \cite{ngpu}, which can learn to multiply long binary numbers.
 %and generalize to input sequences much longer, than the ones seen during training.
 
 %On the other hand, both of these models% use only ``local'' memory access and
 %require at least $\Theta(n^2)$ operations to copy a sequence of length $n$.

%% file: model.tex
\section{Hierarchical Attentive Memory}\label{sec:model}

In this section we describe our novel memory module called Hierarchical Attentive Memory (HAM).
The HAM module is generic and can be used as a building block of larger
neural network architectures. For instance, it can be added to feedforward or LSTM networks
to extend their capabilities.
To make our description more concrete we will consider
a model consisting of an LSTM ``controller'' extended with a HAM module.

The high-level idea behind the HAM module is as follows.
The memory is structured as a full binary tree with the leaves containing the data
stored in the memory.
The inner nodes contain some auxiliary data, which allows us to efficiently
perform some types of ``queries'' on the memory.
In order to access the memory, one starts from the root of the tree and performs
a top-down descent in the tree, which is similar
to the hierarchical softmax procedure \cite{morin2005hierarchical}.
At every node of the tree, one decides to go left or right based on the
auxiliary data stored in this node and a ``query''.
Details are provided in the rest of this section.

\subsection{Notation}

The model takes as input a sequence $x_1,x_2,\ldots$ and outputs a sequence $y_1,y_2,\ldots$.
We assume that each element of these sequences is a binary vector of size $b \in \mathbb{N}$, i.e. $x_i,y_i \in \{0,1\}^b$.
Suppose for a moment that we only want to process input sequences of length $\le n$, where
$n \in \NN$
is a power of two (we show later how to process sequences of an arbitrary length).
The model is based on the full binary tree with $n$ leaves.
Let $V$ denote the set of the nodes in that tree (notice that $|V|=2n-1$)
and let $L \subset V$ denote the set of its leaves.
Let $l(e)$ for $e \in V \setminus L$ be the left child of the node $e$ and let $r(e)$ be its right child.

We will now present the inference procedure for the model and then discuss
how to train it.

\subsection{Inference}\label{section:inference}

\usetikzlibrary{shapes.geometric}
\usetikzlibrary{positioning}
\begin{figure}[t]
\begin{center}
\begin{tikzpicture}[minimum size=0.9cm,scale=0.93, every node/.style={scale=0.7}, triangle/.style = { regular polygon, regular polygon sides=3, rotate=00}]
\node[draw,circle] (y1) at (0,1.5) {$y_1$};
\node[draw,rectangle] (LSTM1) at (0,0) {LSTM};
a\node[draw,triangle,scale=0.4] (HAM1) at (0,-2) {\Huge HAM};

\draw[arrows=-latex] (LSTM1)  to[bend left=0] (y1);
\draw[arrows=-latex] (HAM1)  to[bend left=0] (LSTM1);

\node[draw,circle] (x1) at (-0.7,-3.5) {$x_1$};
\draw[arrows=-latex] (x1)  to[bend left=0] ([yshift=6.5mm]x1.north);
\node () at (0,-3.5) {$\ldots$};
\node[draw,circle] (x2) at (0.7,-3.5) {$x_m$};
\draw[arrows=-latex] (x2)  to[bend left=0] ([yshift=6.5mm]x2.north);

\node[draw,circle] (y2) at (2.5,1.5) {$y_2$};
\node[draw,rectangle] (LSTM2) at (2.5,0) {LSTM};
\node[draw,triangle,scale=0.4] (HAM2) at (2.5,-2) {\Huge HAM};

\draw[arrows=-latex] (LSTM2)  to[bend left=0] (y2);
\draw[arrows=-latex] (HAM2)  to[bend left=0] (LSTM2);

\draw[arrows=-latex] (LSTM1)  to[bend left=0] (LSTM2);
\draw[arrows=-latex] (HAM1)  to[bend left=0] (HAM2);
\draw[arrows=-latex] (LSTM1)  to[bend left=0] (HAM2);

\node[draw,circle] (y3) at (5,1.5) {$y_3$};
\node[draw,rectangle] (LSTM3) at (5,0) {LSTM};
\node[draw,triangle,scale=0.4] (HAM3) at (5,-2) {\Huge HAM};

\draw[arrows=-latex] (LSTM3)  to[bend left=0] (y3);
\draw[arrows=-latex] (HAM3)  to[bend left=0] (LSTM3);

\draw[arrows=-latex] (LSTM2)  to[bend left=0] (LSTM3);
\draw[arrows=-latex] (HAM2)  to[bend left=0] (HAM3);
\draw[arrows=-latex] (LSTM2)  to[bend left=0] (HAM3);

\draw[arrows=-latex] (LSTM3)  to[bend left=0] ([xshift=15mm]LSTM3);
\draw[arrows=-latex] (HAM3)  to[bend left=0] ([xshift=15mm]HAM3);
\node () at (6,-1) {$\ldots$};

\end{tikzpicture}
\end{center}
\caption{The LSTM+HAM model consists of an LSTM controller and a HAM module.
The execution of the model starts with the initialization of HAM using the \emph{whole} input sequence $x_1,x_2,\ldots,x_m$.
At each timestep, the HAM module produces an input for the LSTM, which then produces an output symbol $y_t$.
Afterwards, the hidden states of the LSTM and HAM are updated.
}\label{fig:highlevel}
\end{figure}
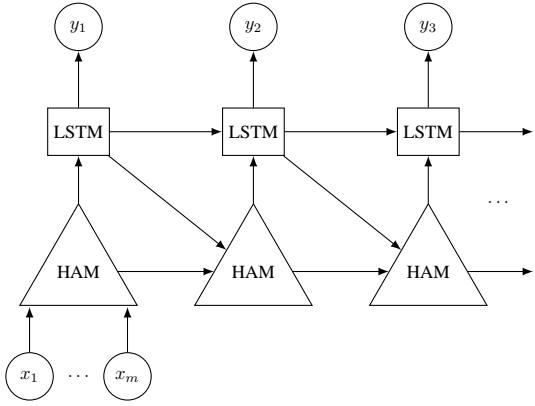

The high-level view of the model execution is presented in Fig.~\ref{fig:highlevel}.
The hidden state of the model consists of two components: the hidden state of the LSTM controller (denoted $h_\texttt{LSTM} \in \RR^l$ for some $l \in \NN$)
and the hidden values stored in the nodes of the HAM tree.
More precisely, for every node $e \in V$ there is a hidden value $h_e \in \RR^d$.
These values change during the recurrent execution of the model, but we drop all timestep indices to simplify the notation.

The parameters of the model describe the input-output behaviour of the LSTM, as well as the following $4$ transformations, which describe the HAM module:
$\emb : \RR^b \rightarrow \RR^d$,
$\join : \RR^d \times \RR^d \rightarrow \RR^d$,
$\search : \RR^d \times \RR^l \rightarrow [0,1]$ and
$\modify : \RR^d \times \RR^l \rightarrow \RR^d$.
These transformations may be represented by arbitrary function approximators, e.g. Multilayer Perceptrons (MLPs).
Their meaning will be described soon.

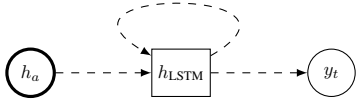
\begin{figure}[h!]
\begin{center}
\begin{tikzpicture}[minimum size=0.9cm,scale=1.0, every node/.style={scale=0.7}]
\node[draw,circle,very thick] (v13) at (0,-1) {$h_a$};
\node[draw,rectangle] (LSTM) at (2,-1) {$h_\text{LSTM}$};
\draw[arrows=-latex,dashed] (v13)  to[bend left=0] (LSTM);
\draw[arrows=-latex,dashed] (LSTM)  to[in=150, out=30, looseness=6] (LSTM);
\node[draw,circle] (y) at (4,-1) {$y_t$};
\draw[arrows=-latex,dashed] (LSTM)  to[bend left=-0] (y);
\end{tikzpicture}
\end{center}
\caption{\emph{Output} phase.
The value $h_a$ stored in the attended leaf is given to the LSTM as an input.
Then, the LSTM produces an output symbol $y_t \in \{0,1\}^b$.
More precisely, the value $u \in \RR^b$ is computed by a trainable linear transformation from $h_\text{LSTM}$
and the distribution of $y_t$ is defined
by the formula
$p(y_{t,i}=1)=\mathbf{sigmoid}(u_i)$ for $1 \le i \le b$.
It may be beneficial to allow the model to access the memory a few times between producing each output symbols.
Therefore, the model produces an output symbol
only at timesteps with indices divisible
by some constant $\eta \in \NN$, which
is a hyperparameter.
}\label{fig:output}
\end{figure}

The details of the model are presented in $4$ figures.
Fig.~\ref{fig:init} describes the initialization of the model.
Each recurrent timestep of the model consists of three phases:
the \emph{attention} phase described in Fig.~\ref{fig:search},
the \emph{output} phase described in Fig.~\ref{fig:output}
and the \emph{update} phase described in Fig.~\ref{fig:write}.
The whole timestep can be performed in time $\Theta(\log n)$.

\newcommand{\tree}[7]{

\node[draw,circle,#1] (v1) at (3.5,3) {$h_1$};

\node[draw,circle] (v2) at (1.5,2) {$h_2$};
\node[draw,circle,#1] (v3) at (5.5,2) {$h_3$};  
  
\node[draw,circle] (v4) at (0.5,1) {$h_4$};
\node[draw,circle] (v5) at (2.5,1) {$h_5$};
\node[draw,circle,#1] (v6) at (4.5,1) {$h_6$};
\node[draw,circle] (v7) at (6.5,1) {$h_7$};  
  
\node[draw,circle] (v8) at (0,0) {$h_8$};
\node[draw,circle] (v9) at (1,0) {$h_9$};
\node[draw,circle] (v10) at (2,0) {$h_{10}$};
\node[draw,circle] (v11) at (3,0) {$h_{11}$};
\node[draw,circle] (v12) at (4,0) {$h_{12}$};
\node[draw,circle,#7] (v13) at (5,0) {$h_{#3}$};
\node[draw,circle] (v14) at (6,0) {$h_{14}$};
\node[draw,circle] (v15) at (7,0) {$h_{15}$};

\draw[arrows=#5,#6] (v1)  to (v2);
\draw[arrows=#5,#4,#6] (v1)  to (v3);
\draw[arrows=#2] (v2)  to (v4);
\draw[arrows=#2] (v2)  to (v5);
\draw[arrows=#5,#4,#6] (v3)  to (v6);
\draw[arrows=#5,#6] (v3)  to (v7);
\draw[arrows=#2] (v4)  to (v8);
\draw[arrows=#2] (v4)  to (v9);
\draw[arrows=#2] (v5)  to (v10);
\draw[arrows=#2] (v5)  to (v11);
\draw[arrows=#5,#6] (v6)  to (v12);
\draw[arrows=#5,#4,#6] (v6)  to (v13);
\draw[arrows=#2] (v7)  to (v14);
\draw[arrows=#2] (v7)  to (v15);
}

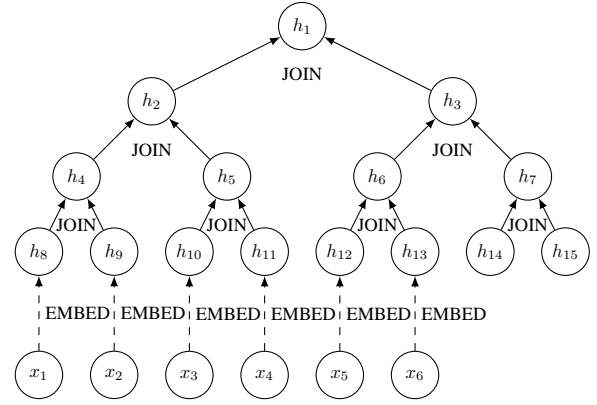
\begin{figure}[h]
\begin{center}
\begin{tikzpicture}[minimum size=0.9cm,scale=1.0, every node/.style={scale=0.7}]

  %\node[above left=-0.1cm and -2.0cm of R] {registers};
  %\node[above left=0.0cm and  -4.7cm of R, text width=2cm,align=center] {outputs of previous modules};
\tree{draw}{latex-}{13}{draw}{latex-}{draw}{draw}

\node[draw,circle,below = 1cm of v8] (x1) {$x_1$};
\node[draw,circle,below = 1cm of v9] (x2) {$x_2$};
\node[draw,circle,below = 1cm of v10] (x3) {$x_3$};
\node[draw,circle,below = 1cm of v11] (x4) {$x_4$};
\node[draw,circle,below = 1cm of v12] (x5) {$x_5$};
\node[draw,circle,below = 1cm of v13] (x6) {$x_6$};

\draw[arrows=-latex,dashed] (x1) -- (v8) node[midway,right] {\emb};
\draw[arrows=-latex,dashed] (x2) -- (v9) node[midway,right] {\emb};
\draw[arrows=-latex,dashed] (x3) -- (v10) node[midway,right] {\emb};
\draw[arrows=-latex,dashed] (x4) -- (v11) node[midway,right] {\emb};
\draw[arrows=-latex,dashed] (x5) -- (v12) node[midway,right] {\emb};
\draw[arrows=-latex,dashed] (x6) -- (v13) node[midway,right] {\emb};

\node[below=0.0cm of v1] {\join};
\node[below=0.0cm of v2] {\join};
\node[below=0.0cm of v3] {\join};
\node[below=0.0cm of v4] {\join};
\node[below=0.0cm of v5] {\join};
\node[below=0.0cm of v6] {\join};
\node[below=0.0cm of v7] {\join};

\end{tikzpicture}
\end{center}
\caption{Initialization of the model.
The value in the $i$-th leaf of HAM is initialized with $\emb(x_i)$, where $\emb$
is a trainable feed-forward network.
If there are more leaves than input symbols, we initialize the values in the excessive leaves with zeros.
Then, we initialize the values in the inner nodes bottom-up
using the formula $h_e = \text{JOIN}(h_{l(e)}, h_{r(e)})$.
The hidden state of the LSTM --- $h_\texttt{LSTM}$ is initialized with zeros.
}\label{fig:init}
\end{figure}

 %\item Computing the query $q \in \RR^d$.
 %\item \emph{Attention} phase: described in Fig.~\ref{fig:search}.
 %\item \emph{Output} phase: described in Fig.~\ref{fig:output}.
 %This step can be skipped in some timesteps, e.g. the model can produce an output symbol every two steps.
 %\item \emph{Write} phase: described in Fig.~\ref{fig:write}.
 %\item \emph{Update} phase: described in Fig.~\ref{fig:update}.

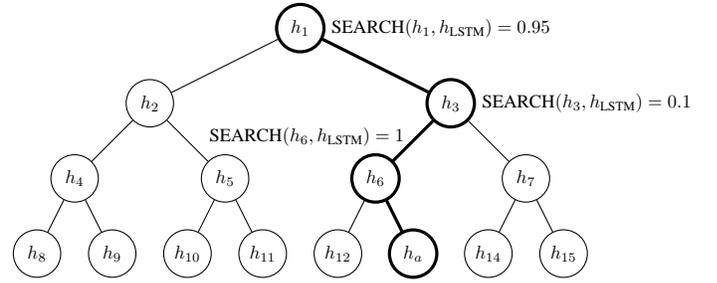
\begin{figure}[h!]
\begin{center}
\begin{tikzpicture}[minimum size=0.9cm,scale=1.0, every node/.style={scale=0.7}]
\tree{very thick}{-}{a}{very thick}{-}{draw}{very thick}
\node[right=0.0cm of v1] {$\search(h_1,h_\text{LSTM})=0.95$};
\node[right=0.0cm of v3] {$\search(h_3,h_\text{LSTM}) = 0.1$};
\node[above left=-0.0cm and -0.7cm of v6] {$\search(h_6,h_\text{LSTM})=1$};
\end{tikzpicture}
\end{center}
\caption{\emph{Attention} phase.
In this phase the model performs a top-down ``search'' in the tree starting from the root.
Suppose that we are currently at the node $c \in V \setminus L$.
We compute the value $p=\text{SEARCH}(h_c, h_\text{LSTM})$.
Then, with probability $p$ the model goes right (i.e. $c := r(c)$)
and with probability $1-p$ it goes left (i.e. $c := l(c)$).
This procedure is continued until we reach one of the leaves.
This leaf is called the \emph{attended} or \emph{accessed} leaf and denoted $a$.
}\label{fig:search}
\end{figure}

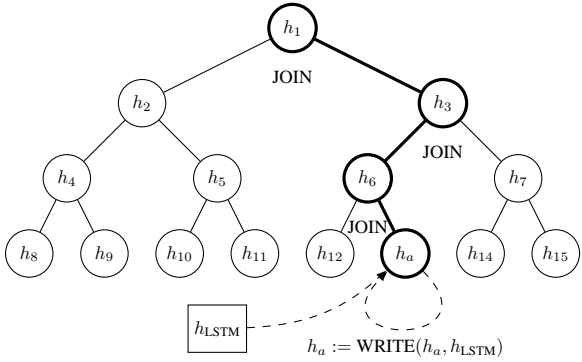
\begin{figure}[h!]
\begin{center}
\begin{tikzpicture}[minimum size=0.9cm,scale=1.0, every node/.style={scale=0.7}]
\tree{very thick}{-}{a}{very thick}{-}{draw}{very thick}
\node[draw,rectangle] (LSTM) at (2.5,-1) {$h_\text{LSTM}$};
\draw[arrows=-latex,dashed] (v13)  to[in=225, out=315, looseness=8] (v13);
\draw[arrows=-latex,dashed] (LSTM)  to[bend right=20] (v13);
\node[below=0.6cm of v13] {$h_a := \modify(h_a,h_\text{LSTM})$};
\node[below=0.0cm of v1] {\join};
\node[below=0.0cm of v3] {\join};
\node[below=0.0cm of v6] {\join};
\end{tikzpicture}
\end{center}
\caption{\emph{Update} phase.
In this phase the value in the attended leaf $a$ is updated.
More precisely, the value is modified using the formula
$h_a := \text{\modify}(h_a, h_\text{LSTM})$.
Then, we update the values of the inner nodes encountered during the attention phase ($h_6,h_3$ and $h_1$ in the figure)
bottom-up using the equation $h_e = \text{JOIN}(h_{l(e)}, h_{r(e)})$.
}\label{fig:write}
\end{figure}

The HAM parameters describe only the $4$ mentioned transformations and hence
the number of the model parameters does not depend on the size of the binary tree used.
Thus, we can use the model to process the inputs of an arbitrary length by using big enough binary trees.
It is not clear that the same set of parameters will give good results across different tree sizes, but we
showed experimentally that it is indeed the case (see Sec.~\ref{sec:exp} for more details).

We decided to represent the transformations defining HAM with
MLPs with ReLU \cite{relu} activation function in all neurons
except the output layer of $\search$, which uses sigmoid activation function to ensure that the output may be interpreted as a probability.
Moreover, the network for $\modify$ is enhanced in a similar way as Highway Networks \cite{highway}, i.e.
$\modify(h_a,h_\text{LSTM}) = T(h_a,h_\text{LSTM}) \cdot H(h_a,h_\text{LSTM}) + \left(1-T(h_a,h_\text{LSTM})\right) \cdot h_a$,
where $H$ and $T$ are two MLPs with sigmoid activation function in the output layer.
This allows the WRITE transformation to easily leave the value $h_a$ unchanged.

\subsection{Training}\label{sec:training}

In this section we describe how to train our model from purely input-output examples
using REINFORCE \cite{reinforce}. In Appendix~\ref{app:soft} we also present a different variant of HAM which
is fully differentiable and can be trained using end-to-end backpropagation.

Let $x,y$ be an input-output pair.
Recall that both $x$ and $y$ are sequences.
Moreover, let $\theta$ denote the parameters of the model and let $A$ denote the sequence of all decisions whether to
go left or right made during \emph{the whole execution} of the model.
We would like to maximize the log-probability of producing the correct output, i.e.
\vspace{-0mm}
$$\mathcal{L} = \log p(y|x,\theta) = \log \left( \sum_A p(A|x,\theta) p(y|A,x,\theta) \right).$$
\vspace{-0mm}

This sum is intractable, so instead of minimizing it directly, we minimize a variational lower bound on it:
\vspace{-0mm}
$$\mathcal{F} =  \sum_A   p(A|x,\theta) \log p(y|A,x,\theta) \le \mathcal{L}.$$
\vspace{-0mm}

This sum is also intractable, so we approximate its gradient using the REINFORCE, which we briefly explain below.
Using the identity $\nabla p(A|x,\theta) = p(A|x,\theta) \nabla \log p(A|x,\theta)$, the gradient of the lower bound with respect to the model parameters can be rewritten as:
\vspace{-2mm}
\begin{equation}
\begin{split}
\nabla \mathcal{F} = \sum_A  p(A|x,\theta) \Big[&\nabla \log p(y|A,x,\theta) \quad + \\
 &\log p(y|A,x,\theta) \nabla \log p(A|x,\theta) \Big]
\end{split}
\end{equation}
\vspace{-5mm}

We estimate this value using Monte Carlo approximation.
For every $x$ we sample $\widetilde{A}$ from $p(A|x,\theta)$ and approximate the gradient for the input $x$ as
$\nabla \log p(y|\widetilde{A},x,\theta) + \log p(y|\widetilde{A},x,\theta) \nabla \log p(\widetilde{A}|x,\theta)$.

Notice that this gradient estimate can be computed using normal backpropagation if we substitute the gradients
in the nodes\footnote{
For a general discussion of computing gradients in computation graphs, which contain stochastic nodes see \cite{schulman2015gradient}.} which sample whether we should go left or right during the \emph{attention} phase by

$$\underbrace{\log p(y|\widetilde{A},x,\theta)}_\text{return} \nabla \log p(\widetilde{A}|x,\theta).$$

This term is called REINFORCE gradient estimate and the left factor is called a \emph{return} in Reinforcement Learning literature.
This gradient estimator is unbiased, but it often has a high variance.
Therefore, we employ two standard variance-reduction technique for REINFORCE: \emph{discounted returns} and \emph{baselines} \cite{reinforce}.
Discounted returns means that our return at the $t$-th timestep has the form $\sum_{t \le i} \gamma^{i-t} \log p(y_i|\widetilde{A},x,\theta)$
for some discount constant $\gamma \in [0,1]$, which is a hyperparameter.
This biases the estimator if $\gamma<1$, but it often decreases its variance.

For the lack of space we do not describe the \emph{baselines} technique.
We only mention that our baseline is case and timestep dependent:
it is computed using a learnable linear transformation from $h_\text{LSTM}$
and trained using MSE loss function. %\todo{Alex suggested that we should describe the baselines.}

The whole model is trained with the Adam \cite{adam} algorithm.
We also employ the following three training techniques:

\paragraph{Different reward function} During our experiments we noticed that better results may be obtained by using a different
reward function for REINFORCE. More precisely,
instead of the log-probability of producing the correct output, we use the percentage of the output bits, which
have the probability of being predicted correctly (given $\widetilde{A}$) greater than $50\%$, i.e.
our discounted return is equal
$\sum_{t \le i,1 \le j \le b}  \gamma^{i-t} \left[p(y_{i,j}|\widetilde{A},x,\theta) >0.5 \right] $.
Notice that it corresponds to the Hamming distance between the most probable outcome
accordingly to the model (given $\widehat{A}$) and the correct output.

\paragraph{Entropy bonus term} We add a special term to the cost function which encourages exploration.
More precisely, for each sampling node we add to the cost function the term
$\frac{\alpha}{H(p)}$, where $H(p)$ is the entropy of the distribution of the decision, whether to go left or right
in this node and $\alpha$ is an exponentially decaying coefficient. This term goes to infinity, whenever the entropy goes to zero, what ensures some level of exploration.
We noticed that this term works better in our experiments than the standard term of the form $-\alpha H(p)$ \cite{reinforce}.

\paragraph{Curriculum schedule}
We start with training on inputs with lengths sampled
uniformly from $[1,n]$ for some $n=2^k$ and the binary tree with $n$
leaves.
Whenever the error drops below some threshold, we
increment the value $k$ and start using the bigger tree with $2n$ leaves
and inputs with lengths sampled uniformly from $[1,2n]$.

%% file: experiments.tex
\section{Experiments}\label{sec:exp}

In this section, we evaluate two variants of using the HAM module. The
first one is the model described in Sec.~\ref{sec:model}, which combines an LSTM controller
with a HAM module (denoted by LSTM+HAM).
Then, in Sec.~\ref{sec:ham}
we investigate the ``raw'' HAM (without the LSTM controller) to check its
capability of acting as classic data structures: a stack, a FIFO queue and
a priority queue.

\subsection{Test setup}\label{sec:test_setup}

For each test that we perform, we apply the following procedure. 
First, we train the model with memory of size up to $n=32$ using the curriculum schedule described in Sec.~\ref{sec:training}.
The model is trained using the minibatch Adam algorithm with exponentially decaying learning rate.
We use random search to determine the best hyper-parameters for the model.
We use gradient clipping \cite{pascanu2012understanding} with constant $5$.
The depth of our MLPs is either $1$ or $2$,
the LSTM controller has $l=20$ memory cells and the hidden values in the tree have dimensionality $d=20$.
Constant $\eta$ determining a number of memory accesses between producing each output symbols (Fig.~\ref{fig:output})
is equal either $1$ or $2$.
We always train for $100$ epochs, each consisting of $1000$ batches of size $50$.
After each epoch we evaluate the model on $200$ validation batches without learning.
When the training is finished, we select the model parameters that gave the lowest error rate on validation batches
and report the error using these parameters on fresh $2,500$ random examples.

We report two types of errors: a test error and a generalization error.
The test error shows how well the model is able to fit the data distribution and generalize 
to unknown cases, assuming that cases of similar lengths were shown during the training.
It is computed using the HAM memory with $n=32$ leaves, as the percentage of output \emph{sequences}, which were 
predicted incorrectly.
The lengths of test examples are sampled uniformly from the range $[1, n]$.
Notice that we mark the whole output sequence as incorrect even
if only one bit was predicted incorrectly, e.g.
a hypothetical model predicting each bit incorrectly with probability $1\%$
(and independently of the errors on the other bits) has an error rate of $96\%$
on \emph{whole sequences}
if outputs consist of $320$ bits.

The generalization error shows how well the model performs with enlarged memory on examples 
with lengths exceeding $n$. We test our model with memory $4$ times bigger than the training
one. The lengths of input sequences are now sampled uniformly from the range $[2n+1, 4n]$.

During testing we make our model fully deterministic by using the most probable
outcomes instead of stochastic sampling.  More precisely, we assume that during
the \emph{attention phase} the model decides to go right iff $p>0.5$
(Fig.~\ref{fig:search}).  Moreover, the output symbols (Fig.~\ref{fig:output})
are computed by rounding to zero or one instead of sampling.

\subsection{LSTM+HAM}\label{sec:lstm+ham}

We evaluate the model on a number of algorithmic tasks described below:

  %\item
  \paragraph{\texttt{Reverse}:} Given a sequence of $10$-bit vectors, output them in the reversed order., i.e.
  $y_i = x_{m+1-i}$ for $1 \le i \le m$, where $m$ is the length of the input sequence.
  %\item
  \paragraph{\texttt{Search}:} Given a sequence of pairs $x_i = \textbf{key}_i || \textbf{value}_i$
  for $1 \le i \le m-1$ sorted by keys and a query $x_m=q$, find
  the smallest $i$ such that $\textbf{key}_i=q$ and output $y_1=\textbf{value}_i$.
  Keys and values are $5$-bit vectors and keys are compared lexicographically.
  The LSTM+HAM model is given only two timesteps ($\eta=2$) to solve this problem, which
  forces it to use a form of binary search.
  %\item
  \paragraph{\texttt{Merge}:} Given two \emph{sorted} sequences of pairs --- $(p_1,v_1),\ldots,(p_m,v_m)$ and $(p_1',v_1'),\ldots,(p_{m'}',v_{m'}')$, where
  $p_i,p_i' \in [0,1]$ and $v_i,v_i' \in \{0,1\}^5$, merge them.
  Pairs are compared accordingly to their priorities, i.e. values $p_i$ and $p_i'$.
  Priorities are unique and sampled uniformly from the set $\{\frac{1}{300},\ldots,\frac{300}{300}\}$, because
  neural networks can not easily distinguish two real numbers which are very close to each other.
  Input is encoded as $x_i = p_i || v_i$ for $1 \le i \le m$
  and $x_{m+i} = p_i' || v_i'$ for $1 \le i \le m'$.
  The output consists of the vectors $v_i$ and $v_i'$ sorted accordingly to their priorities\footnote{
  Notice that we earlier assumed for the sake of simplicity that the input sequences consist of \emph{binary} vectors
  and in this task the priorities are \emph{real} values.
  It does not however require any change of our model.
  We decided to use real priorities in this task in order to diversify our set of problems.}.
  %\item
  \paragraph{\texttt{Sort}:} Given a sequence of pairs $x_i = \textbf{key}_i || \textbf{value}_i$ sort them in a stable way\footnote{Stability means that pairs with equal keys should
  be ordered accordingly to their order in the input sequence.}
  accordingly to the lexicographic order of the keys.
  Keys and values are $5$-bit vectors.
  %\item
  \paragraph{\texttt{Add}:} Given two numbers represented in binary, compute their sum. The input is represented as
  $a_1,\ldots,a_m,\textbf{+},b_1,\ldots,b_m,\textbf{=}$ (i.e. $x_1=a_1,x_2=a_2$ and so on), where $a_1,\ldots,a_m$ and $b_1,\ldots,b_m$
  are bits of the input numbers and $\textbf{+},\textbf{=}$ are some special symbols.
  Input and output numbers are encoded
  starting from the \emph{least} significant bits.

Every example output shown during the training is finished by a special 
``End Of Output'' symbol, which the model learns to predict.
It forces the model to learn not only the output symbols, but
also the length of the correct output.

We compare our model with 2 strong baseline models: encoder-decoder LSTM
\cite{sutskever2014sequence} and encoder-decoder LSTM with attention (denoted
LSTM+A) \cite{bahdanau2014}.  
The number of the LSTM cells in the baselines was
chosen in such a way, that they have more parameters than the biggest of our
models. We also use random search to select an optimal learning rate and some other
parameters for the baselines and train them using the same curriculum scheme as LSTM+HAM.

The results are presented in Table~\ref{fig:model_results}.
Not only, does LSTM+HAM solve all the problems almost perfectly, but it also generalizes
very well to much longer inputs on all problems except \texttt{Add}.
Recall that for the generalization tests we used a HAM memory of a different size than the ones
used during the training, what shows that HAM generalizes very well to new sizes of the binary tree.
We find this fact quite interesting, because it means that
parameters learned from a small neural network (i.e. HAM based on a tree with $32$ leaves)
can be successfully used in a different, bigger network (i.e. HAM with $128$ memory cells).

In comparison, the LSTM with attention does not learn to merge, nor sort.
It also completely fails to generalize to longer examples, which shows
that LSTM+A learns rather some statistical dependencies between inputs and outputs than
the real algorithms.%, which should work for inputs of arbitrary lengths.

The LSTM+HAM model makes a few errors when testing on longer outputs than the ones encountered during the training.
Notice however, that we show in the table the percentage of output sequences, which
contain \emph{at least one} incorrect bit.
For instance, LSTM+HAM on the problem \texttt{Merge}
predicts incorrectly only $0.03$\% of output bits, which corresponds to $2.48\%$ of
incorrect output sequences.
We believe that these rare mistakes could be avoided if one trained the model longer
and chose carefully the learning rate schedule.
One more way to boost generalization capabilities would be to simultaneously train the models
with different memory sizes and shared parameters. We have not tried this as the generalization properties of the model
were already very good.

\newcolumntype{P}[1]{>{\centering\arraybackslash}p{#1}}

\begin{table}[h]\small \caption{Experimental results.
The upper table presents the error rates on inputs
of the same lengths as the ones used during training.
The lower table shows the error rates on input sequences
$2$ to $4$ times longer than the ones encountered during training.
LSTM+A denotes an LSTM with the standard attention mechanism.
Each error rate is a percentage of \emph{output sequences}, which contained at least
one incorrectly predicted bit.
}

\begin{center}
\begin{tabular}{|c|c|c|c|}%P{10mm}|}
\hline
\textbf{test error} & LSTM & LSTM+A & \textbf{LSTM+HAM} \\ % & LSTM+ \newline DHAM \\
\hline
\texttt{Reverse} & $73$\% & $0$\% & $\mathbf{0}$\textbf{\%} \\% & $0$\% \\
\hline
\texttt{Search} & $62$\% & $0.04$\% & $\mathbf{0.12}$\textbf{\%} \\% & $0.3$\% \\
\hline
\texttt{Merge} & $88$\% & $16$\% & $\mathbf{0}$\textbf{\%} \\% & ? \\
\hline
\texttt{Sort} & $99$\% & $25$\% & $\mathbf{0.04}$\textbf{\%} \\% & ? \\
\hline
\texttt{Add} & $39$\% & $0$\% &  $\mathbf{0}$\textbf{\%} \\% & ? \\
\hline
\hline
\textbf{2-4x longer inputs} & LSTM & LSTM+ \newline A & \textbf{LSTM+}\newline \textbf{HAM} \\ % & LSTM+ \newline DHAM \\
\hline
\texttt{Reverse} & $100$\% & $100$\% & $\mathbf{0}$\textbf{\%} \\% & ? \\
\hline
\texttt{Search} & $89$\% & $0.52$\% & $\mathbf{1.68}$\textbf{\%} \\% & ? \\
\hline
\texttt{Merge} & $100$\% & $100$\% & $\mathbf{2.48}$\textbf{\%} \\% & ? \\
\hline
\texttt{Sort} & $100$\% & $100$\% & $\mathbf{0.24}$\textbf{\%} \\% & ? \\
\hline
\texttt{Add} & $100$\% & $100$\% & $\mathbf{100}$\textbf{\%} \\% & ? \\
\hline
\hline
\textbf{Complexity} & $\Theta(1)$ & $\Theta(n)$ & $\mathbf{\Theta(\textbf{log}~n)}$ \\% & $\Theta(n)$\\
\hline
\end{tabular}
\end{center}
\label{fig:model_results}
\end{table}

\subsection{Raw HAM}\label{sec:ham}

In this section, we evaluate ``raw'' HAM module (without the LSTM controller) to
see if it can act as a drop-in replacement for $3$ classic data structures: a stack, 
a FIFO queue and a priority queue.
For each task, the network is given a sequence of PUSH and POP
operations in an \emph{online} manner: at timestep $t$ the network sees only the
$t$-th operation to perform $x_t$.
This is a more realistic scenario for data structures
usage as it prevents the network from cheating by peeking into the future.

Raw HAM module differs from the LSTM+HAM model from Sec.~\ref{sec:model}
in the following way:
\begin{itemize}
\item The HAM memory is initialized with zeros.

\item The $t$-th output symbol $y_t$ is computed using an MLP
from the value in the accessed leaf $h_a$.% at the timestep $t$.

\item Notice that in the LSTM+HAM model, $h_\text{LSTM}$
acted as a kind of ``query'' or ``command'' guiding the behaviour of HAM.
We will now use the values $x_t$ instead.
Therefore,
at the $t$-th timestep we use $x_t$ instead of $h_\text{LSTM}$
whenever $h_\text{LSTM}$ was used in the original model, e.g. during the \emph{attention} phase
(Fig.~\ref{fig:search}) we use $p=\text{SEARCH}(h_c, x_t)$ instead of
$p=\text{SEARCH}(h_c, h_\text{LSTM})$.

\end{itemize}

We evaluate raw HAM on the following tasks: 
 %\texttt{Stack},
 %\texttt{Queue} and \texttt{PriorityQueue}.
 %In each case input symbols are 
 %\begin{enumerate}
 %\item
 \paragraph{\texttt{Stack}:} The ``PUSH $x$'' operation places the element $x$ (a $5$-bit vector) on top of
 the stack, and the ``POP'' returns the last added element and removes it from the stack.
 %\item
 \paragraph{\texttt{Queue}:} The ``PUSH $x$'' operation places the element $x$ (a $5$-bit vector) at the
 end of the queue and the ``POP'' returns the oldest element and removes it from the queue.
 \paragraph{\texttt{PriorityQueue}:} The ``PUSH $x$ $p$'' operations adds the element
 $x$ with priority $p$ to the queue. The ``POP'' operation returns the value with
 the highest priority and remove it from the queue. Both $x$ and $p$ are
 represented as $5$-bit vectors and priorities are compared lexicographically.
 To avoid ties we assume that all elements have different priorities.
 %For two elements $e_1 = (x_1, p_1)$ and $e_2 = (x_2, p_2)$
 %we say that $e_1$ has lower priority than $e_2$ iff $p_1$ is
 %lexicographically smaller than $p_2$.
 %\end{enumerate}

Model was trained with the memory of size up to $n=32$ with operation sequences of length $n$.
Sequences of PUSH/POP actions for training were selected randomly.
The $t$-th operation out of $n$ operations in the sequence was POP with probability $\frac{t}{n}$ and PUSH otherwise.
To test generalization, we report the error rates with the memory of size $4n$
on sequences of operations of length $4n$.

The results presented in Table \ref{fig:ds_results} shows that HAM simulates a stack and a queue perfectly
with no errors whatsoever even for memory $4$ times bigger.
For the \texttt{PriorityQueue} task, the model generalizes almost perfectly to large memory,
with errors only in $0.2$\% of output sequences.

\begin{table}[h]\small
\caption{Results of experiments with the raw version of HAM (without the LSTM controller).
Error rates are measured as a percentage of operation sequences in which \emph{at least one} POP query was
not answered correctly.
}
\begin{center}
\begin{tabular}{|c|P{21mm}|P{21mm}|}
\hline
\textbf{Task} & \textbf{Test Error} &  \textbf{Generalization Error} \\
\hline
\texttt{Stack} & $0$\% &  $0$\%  \\   % tree161, task101, local eval generalization
\hline
\texttt{Queue} & $0$\% & $0$\% \\  % tree161, task201, local eval generalization
\hline
\texttt{PriorityQueue} & $0.08$\% & $0.2$\% \\  % tree166, task272
\hline
\end{tabular}
\end{center}
\label{fig:ds_results}
\end{table}

\subsection{Analysis}

In this section, we present some insights into the algorithms learned by the LSTM+HAM model,
by investigating the the hidden representations $h_e$ learned for a variant of the problem \texttt{Sort}
in which we sort $4$-bit vectors lexicographically\footnote{
In the problem \texttt{Sort} considered in the experimental results, there are separate keys and values, which
forces the model to learn stable sorting.
Here, for the sake of simplicity, we consider the simplified version of the problem and do not use separate keys and values.}.
For demonstration purposes, we use a small tree with $n=8$ leaves and $d=6$.

The trained network performs sorting perfectly.
It attends to the leaves in the order corresponding to the order
of the sorted input values, i.e.
at every timestep HAM attends to the
leaf corresponding to the smallest input value among
the leaves, which have not been attended so far.

It would be interesting to exactly understand the algorithm used by the network to
perform this operation.
A natural solution to this problem would be to store in each hidden node $e$ the
smallest input value among the (unattended so far) leaves \emph{below} $e$ together with the information whether
the smallest value is in the right or the left subtree under $e$.

We present two timesteps of our model together with some insights into the
algorithm used by the network in Fig.\ref{fig:example}.

\definecolor{Gray}{gray}{0.8}
\newcolumntype{g}{>{\columncolor{Gray}}c}

\newcommand{\ivalues}[6]{
\definecolor{c1}{rgb}{#3,#3,1}
\definecolor{c2}{rgb}{#4,#4,1}
\definecolor{c3}{rgb}{#5,#5,1}
\definecolor{c4}{rgb}{#6,#6,1}
\node[box,fill=c1] at (#1,#2-1){};
\node[box,fill=c2] at (#1,#2-1-0.16*1){};
\node[box,fill=c3] at (#1,#2-1-0.16*2){};
\node[box,fill=c4] at (#1,#2-1-0.16*3){};

\draw[arrows=-latex,dashed] (#1,#2-1+0.08) -- (#1,#2-0.3);
}

\newcommand{\invvalues}[6]{
\node[box, opacity=0] at (#1,#2-1){};
\node[box, opacity=0] at (#1,#2-1-0.16*1){};
\node[box, opacity=0] at (#1,#2-1-0.16*2){};
\node[box, opacity=0] at (#1,#2-1-0.16*3){};
}

\newcommand{\values}[8]{
\definecolor{c1}{rgb}{#3,#3,1}
\definecolor{c2}{rgb}{#4,#4,1}
\definecolor{c3}{rgb}{#5,#5,1}
\definecolor{c4}{rgb}{#6,#6,1}
\definecolor{c5}{rgb}{#7,#7,1}
\definecolor{c6}{rgb}{1,#8,1}

\node[box,fill=c1] at (#1-0.17,#2+0.08){};
\node[box,fill=c2] at (#1,#2+0.08){};
\node[box,fill=c3] at (#1+0.17,#2+0.08){};
\node[box,fill=c4] at (#1-0.17,#2-0.08){};
\node[box,fill=c5] at (#1,#2-0.08){};
\node[box,fill=c6] at (#1+0.17,#2-0.08){};
}

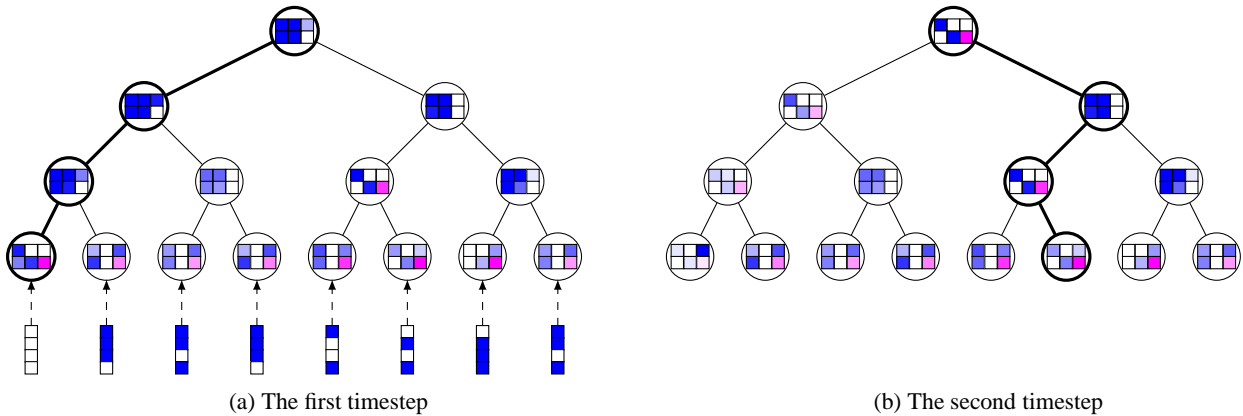
\begin{figure*}[t]
\begin{subfigure}[t]{0.5\textwidth}
\begin{tikzpicture}[minimum size=0.9cm,scale=1.0, every node/.style={scale=0.7},box/.style={rectangle,draw=black, minimum size=0.2}]
\node[draw,circle,very thick] (v1) at (3.5,3) {};

\node[draw,circle,very thick] (v2) at (1.5,2) {};
\node[draw,circle] (v3) at (5.5,2) {};  
  
\node[draw,circle,very thick] (v4) at (0.5,1) {};
\node[draw,circle] (v5) at (2.5,1) {};
\node[draw,circle] (v6) at (4.5,1) {};
\node[draw,circle] (v7) at (6.5,1) {};  
  
\node[draw,circle,very thick] (v8) at (0,0) {};
\node[draw,circle] (v9) at (1,0) {};
\node[draw,circle] (v10) at (2,0) {};
\node[draw,circle] (v11) at (3,0) {};
\node[draw,circle] (v12) at (4,0) {};
\node[draw,circle] (v13) at (5,0) {};
\node[draw,circle] (v14) at (6,0) {};
\node[draw,circle] (v15) at (7,0) {};

\draw[arrows=-,very thick] (v1)  to (v2);
\draw[arrows=-] (v1)  to (v3);
\draw[arrows=-,very thick] (v2)  to (v4);
\draw[arrows=-] (v2)  to (v5);
\draw[arrows=-] (v3)  to (v6);
\draw[arrows=-] (v3)  to (v7);
\draw[arrows=-,very thick] (v4)  to (v8);
\draw[arrows=-] (v4)  to (v9);
\draw[arrows=-] (v5)  to (v10);
\draw[arrows=-] (v5)  to (v11);
\draw[arrows=-] (v6)  to (v12);
\draw[arrows=-] (v6)  to (v13);
\draw[arrows=-] (v7)  to (v14);
\draw[arrows=-] (v7)  to (v15);

\ivalues{0}{0}{1.0}{1.0}{1.0}{1.0}
\ivalues{1}{0}{0.0}{0.0}{0.0}{1.0}
\ivalues{2}{0}{0.0}{0.0}{1.0}{0.0}
\ivalues{3}{0}{0.0}{0.0}{0.0}{1.0}
\ivalues{4}{0}{0.0}{1.0}{1.0}{0.0}
\ivalues{5}{0}{1.0}{0.0}{1.0}{0.0}
\ivalues{6}{0}{1.0}{0.0}{0.0}{0.0}
\ivalues{7}{0}{0.0}{0.0}{1.0}{0.0}
\values{3.5}{3}{0}{0}{0.7}{0}{0}{1.0}
\values{1.5}{2}{0}{0}{0.1}{0}{0}{1.0}
\values{5.5}{2}{0}{0.0}{1.0}{0.1}{0}{1.0}
\values{0.5}{1}{0}{0}{0.5}{0}{0.1}{1.0}
\values{2.5}{1}{0.4}{0.4}{1.0}{0.5}{0.6}{1.0}
\values{4.5}{1}{0}{1.0}{1.0}{1.0}{0.1}{0.2}
\values{6.5}{1}{0}{0}{0.9}{0}{0.4}{1.0}
\values{0}{0}{0.1}{1.0}{1.0}{0.5}{0.2}{0}
\values{1}{0}{0.7}{1.0}{0.3}{0.2}{1.0}{0.5}
\values{2}{0}{0.6}{1.0}{0.4}{0.5}{1.0}{0.6}
\values{3}{0}{0.7}{1.0}{0.3}{0.2}{1.0}{0.5}
\values{4}{0}{0.3}{1.0}{0.5}{0.4}{1.0}{0.2}
\values{5}{0}{0.6}{1.0}{0.8}{1.0}{0.5}{0}
\values{6}{0}{1.0}{1.0}{0.6}{1.0}{0.7}{0}
\values{7}{0}{0.6}{1.0}{0.4}{0.5}{1.0}{0.6}

\end{tikzpicture}
\caption{The first timestep}
\end{subfigure}
~
\begin{subfigure}[t]{0.5\textwidth}
\begin{tikzpicture}[minimum size=0.9cm,scale=1.0, every node/.style={scale=0.7},box/.style={rectangle,draw=black, minimum size=0.2},]
\node[draw,circle,very thick] (v1) at (3.5,3) {};

\node[draw,circle] (v2) at (1.5,2) {};
\node[draw,circle,very thick] (v3) at (5.5,2) {};  
  
\node[draw,circle] (v4) at (0.5,1) {};
\node[draw,circle] (v5) at (2.5,1) {};
\node[draw,circle,very thick] (v6) at (4.5,1) {};
\node[draw,circle] (v7) at (6.5,1) {};  
  
\node[draw,circle] (v8) at (0,0) {};
\node[draw,circle] (v9) at (1,0) {};
\node[draw,circle] (v10) at (2,0) {};
\node[draw,circle] (v11) at (3,0) {};
\node[draw,circle] (v12) at (4,0) {};
\node[draw,circle,very thick] (v13) at (5,0) {};
\node[draw,circle] (v14) at (6,0) {};
\node[draw,circle] (v15) at (7,0) {};

\draw[arrows=-] (v1)  to (v2);
\draw[arrows=-,very thick] (v1)  to (v3);
\draw[arrows=-] (v2)  to (v4);
\draw[arrows=-] (v2)  to (v5);
\draw[arrows=-,very thick] (v3)  to (v6);
\draw[arrows=-] (v3)  to (v7);
\draw[arrows=-] (v4)  to (v8);
\draw[arrows=-] (v4)  to (v9);
\draw[arrows=-] (v5)  to (v10);
\draw[arrows=-] (v5)  to (v11);
\draw[arrows=-] (v6)  to (v12);
\draw[arrows=-,very thick] (v6)  to (v13);
\draw[arrows=-] (v7)  to (v14);
\draw[arrows=-] (v7)  to (v15);
\values{3.5}{3}{0}{1.0}{1.0}{1.0}{0}{0}
\values{1.5}{2}{0.3}{1.0}{1.0}{1.0}{0.6}{0.7}
\values{5.5}{2}{0}{0.0}{1.0}{0.1}{0}{1.0}
\values{0.5}{1}{0.8}{0.9}{1.0}{1.0}{0.8}{0.7}
\values{2.5}{1}{0.4}{0.4}{1.0}{0.5}{0.6}{1.0}
\values{4.5}{1}{0}{1.0}{1.0}{1.0}{0.1}{0.2}
\values{6.5}{1}{0}{0}{0.9}{0}{0.4}{1.0}
\values{0}{0}{0.9}{1.0}{0.0}{1.0}{0.9}{0.9}
\values{1}{0}{0.7}{1.0}{0.3}{0.2}{1.0}{0.5}
\values{2}{0}{0.6}{1.0}{0.4}{0.5}{1.0}{0.6}
\values{3}{0}{0.7}{1.0}{0.3}{0.2}{1.0}{0.5}
\values{4}{0}{0.3}{1.0}{0.5}{0.4}{1.0}{0.2}
\values{5}{0}{0.6}{1.0}{0.8}{1.0}{0.5}{0}
\values{6}{0}{1.0}{1.0}{0.6}{1.0}{0.7}{0}
\values{7}{0}{0.6}{1.0}{0.4}{0.5}{1.0}{0.6}

\invvalues{0}{0}{1.0}{1.0}{1.0}{1.0}
\invvalues{1}{0}{0.0}{0.0}{0.0}{1.0}
\invvalues{2}{0}{0.0}{0.0}{1.0}{0.0}
\invvalues{3}{0}{0.0}{0.0}{0.0}{1.0}
\invvalues{4}{0}{0.0}{1.0}{1.0}{0.0}
\invvalues{5}{0}{1.0}{0.0}{1.0}{0.0}
\invvalues{6}{0}{1.0}{0.0}{0.0}{0.0}
\invvalues{7}{0}{0.0}{0.0}{1.0}{0.0}
\end{tikzpicture}
\caption{The second timestep}
\end{subfigure}

\caption{%An exemplary input sequence and the state of HAM after initialization (left) and after
This figure shows two timesteps of the model.
The LSTM controller is not presented to simplify the exposition.
The input sequence is presented on the left, below the tree:
$x_1=\texttt{0000},x_2=\texttt{1110},x_3=\texttt{1101}$ and so on.
The 2x3 grids in the nodes of the tree represent the values $h_e \in \RR^6$.
White cells correspond to value $0$ and non-white cells correspond to values $>0$.
The lower-rightmost cells are presented in pink, because we managed to decipher the meaning of this coordinate for the inner nodes. This coordinate in the node $e$ denotes
whether the minimum in the subtree (among the values unattended so far) is in the right or left subtree of $e$.  Value greater than $0$ (pink in the picture)
means that the minimum is in the right subtree and therefore we should go right
while visiting this node in the \emph{attention} phase.
In the first timestep the leftmost leaf (corresponding to the input \texttt{0000})
is accessed.
Notice that the last coordinates (shown in pink) are updated appropriately,
e.g. the smallest unattended value at the beginning of the second timestep is \texttt{0101}, which corresponds to the $6$-th leaf.
It is in the right subtree under the root and accordingly
the last coordinate in the hidden value stored in the root is high (i.e. pink in the figure).
}\label{fig:example}
\end{figure*}

%% file: comparison.tex
\section{Comparison to other models}

Comparing neural networks able to learn algorithms is difficult for a few reasons.
First of all, there are no well-established benchmark problems for this area.
Secondly, the difficulty of a problem often depends on the way 
inputs and outputs are encoded.
For example, the difficulty of the problem of adding long binary numbers depends on whether the numbers are
\emph{aligned} (i.e. the $i$-th bit of the second number is ``under'' the $i$-th bit of the first number)
or written next to each other (e.g. 10011+10101).
Moreover, we could compare error rates on inputs from the same distribution as the ones
seen during the training or compare error rates on inputs longer than the ones seen during the training to see
if the model ``really learned the algorithm''.
Furthermore, different models scale differently with the memory size, which makes direct comparison of
error rates less meaningful.

As far as we know, our model is the first one which is able to learn a sorting
algorithm from pure input-output examples.
In \cite{npi} it is shown that an LSTM is able to learn to sort short sequences, but it fails
to generalize to inputs longer than the ones seen during the training.  It is quite
clear that an LSTM can not learn a ``real'' sorting algorithm, because it uses
a bounded memory independent of the length of the input.  The Neural
Programmer-Interpreter \cite{npi} is a neural network architecture, which is
able to learn bubble sort, but it requires strong supervision in the form of
execution traces.
In comparison, our model can be trained from pure input-output examples, which
is crucial if we want to use it to solve problems for which we do not know any
algorithms.

An important feature of neural memories is their efficiency.
Our HAM module in comparison to many other recently proposed solutions is effective
and allows to access the memory in $\Theta(\log(n))$ complexity.
In the context of learning algorithms it may sound surprising that
among all the architectures mentioned in Sec.~\ref{sec:related} the only ones, which
can copy a sequence of length $n$ without $\Theta(n^2)$ operations are:
Reinforcement-Learning NTM \cite{rl-ntm},
the model from \cite{zaremba2015learning},
Neural Random-Access Machine \cite{nram}, and
Queue-Augmented LSTM \cite{grefenstette2015learning}.
However, the first three models have been only successful on relatively simple tasks.
The last model was successful on some synthetic tasks from the domain of Natural Language Processing, which
are very different from the tasks we tested our model on, so we can not directly compare the two models.

Finally, we do not claim that our model is superior to the all other ones, e.g. Neural Turing Machines (NTM) \cite{ntm}.
We believe that both memory mechanisms are complementary: NTM memory has a built-in associative map functionality, which
may be difficult to achieve in HAM. On the other hand, HAM
performs better in tasks like sorting due to a built-in bias towards
operating on intervals of memory cells.
Moreover, HAM allows much more efficient memory access than NTM.
It is also quite possible that a machine able to learn algorithms should use many different types of memory in
the same way as human brain stores a piece of information differently depending on
its type and how long it should be stored \cite{berntson2009handbook}.
\vspace{-2mm}

%% file: conclusions.tex
\section{Conclusions}

We presented a new memory architecture for neural networks
called Hierarchical Attentive Memory.
Its crucial property is
that it scales well with the memory size --- the memory access requires only
$\Theta(\log n)$ operations. This complexity is
achieved by using a new attention mechanism based on a binary tree.
The novel attention mechanism is not only
faster than the standard one used in Deep Learning, but it
also facilities learning algorithms due to the embedded tree structure.

We showed that an LSTM augmented with HAM can learn a number of algorithms like merging, sorting
or binary searching % or adding numbers represented in binary
from pure input-output examples.
In particular, it is the first neural architecture able to learn a sorting algorithm and generalize well to
sequences much longer than the ones seen during the training.

We believe that some concepts used in HAM, namely
the novel attention mechanism and the
idea of aggregating information through a binary tree
may find applications in Deep Learning
outside of the problem of designing neural memories.

\subsection*{Acknowledgements}

We would like to thank Nando de Freitas, Alexander Graves, Serkan Cabi, Misha Denil and Jonathan Hunt for helpful comments and discussions.

%% file: soft.tex
\section{Using soft attention}\label{app:soft}

One of the open questions in the area of designing neural networks with
attention mechanisms is whether to use a \emph{soft} or \emph{hard} attention.
The model described in the paper belongs to the latter class of attention
mechanisms as it makes hard, stochastic choices.  The other solution would be
to use a soft, differentiable mechanism, which attends to a linear combination
of the potential attention targets and do not involve any sampling.  The main
advantage of such models is that their gradients can be computed exactly.

We now describe how to modify the model to make it fully
differentiable ("DHAM").  Recall that in the original model the leaf which is attended
at every timestep is sampled stochastically.  Instead of that, we will now at
every timestep compute for every leaf $e$ the probability $p(e)$ that this leaf
would be attended if we used the stochastic procedure described in Fig.~\ref{fig:search}.
The value $p(e)$ can be computed by multiplying the
probabilities of going in the right direction from all the nodes on the path
from the root to $e$.

As the input for the LSTM we then use the value $\sum_{e \in L} p(e) \cdot h_e$.
During the \emph{write} phase, we update the values of \emph{all} the leaves
using the formula
$h_e := p(e) \cdot \modify(h_e, h_\text{ROOT}) + (1-p(e)) \cdot h_e$.
Then, in the \emph{update} phase we update the values of \emph{all} the inner nodes, so that
the equation $h_e = \text{JOIN}(h_{l(e)}, h_{r(e)})$ is satisfied for each inner node $e$.
Notice that one timestep of the soft version of the model takes time
$\Theta(n)$ as we have to update the values of all the nodes in the tree.
Our model may be seen as a special case of Gated Graph Neural Network \cite{li2015gated}.

This version of the model is fully differentiable and therefore it can be trained using end-to-end backpropagation
on the log-probability of producing the correct output. We observed that training DHAM is
slightly easier than the REINFORCE version. However, DHAM does not generalize as well as HAM to larger memory sizes.